\documentclass[a4paper, twocolumn, 11pt, twoside]{article}

\usepackage[english]{babel}
\usepackage{linguamatica}

\usepackage{amsmath}
\usepackage[nolist]{acronym}
\usepackage{framed}
\usepackage{dialogue}

\usepackage{todonotes}

\submitted{\OCT{} 2018}
\accepted{\JUL{} 2019}

\title{Hierarchical Multi-Label Dialog Act Recognition on Spanish Data}

\titleEN{Translated version of the article published in Linguamática~\citep{Ribeiro2019}. \\ Please cite its original version.}

\author{
  Eugénio Ribeiro
  \instituto{L$^2$F {--} Spoken Language Systems Laboratory {--} INESC-ID Lisboa \nl
  Instituto Superior Técnico, Universidade de Lisboa, Portugal}
  \email{eugenio.ribeiro@inesc-id.pt} 
  \and 
  Ricardo Ribeiro
  \instituto{L$^2$F {--} Spoken Language Systems Laboratory {--} INESC-ID Lisboa \nl
  Instituto Universitário de Lisboa (ISCTE-IUL), Portugal}
  \email{ricardo.ribeiro@inesc-id.pt}
  \and
  David Martins de Matos
  \instituto{L$^2$F {--} Spoken Language Systems Laboratory {--} INESC-ID Lisboa \nl
  Instituto Superior Técnico, Universidade de Lisboa, Portugal}
  \email{david.matos@inesc-id.pt} 
}

\begin{document}
\maketitle

%
%

\begin{abstract}

Dialog acts reveal the intention behind the uttered words. Thus, their automatic recognition is important for a dialog system trying to understand its conversational partner. The study presented in this article approaches that task on the DIHANA corpus, whose three-level dialog act annotation scheme poses problems which have not been explored in recent studies. In addition to the hierarchical problem, the two lower levels pose multi-label classification problems. Furthermore, each level in the hierarchy refers to a different aspect concerning the intention of the speaker both in terms of the structure of the dialog and the task. Also, since its dialogs are in Spanish, it allows us to assess whether the state-of-the-art approaches on English data generalize to a different language. More specifically, we compare the performance of different segment representation approaches focusing on both sequences and patterns of words and assess the importance of the dialog history and the relations between the multiple levels of the hierarchy. Concerning the single-label classification problem posed by the top level, we show that the conclusions drawn on English data also hold on Spanish data. Furthermore, we show that the approaches can be adapted to multi-label scenarios. Finally, by hierarchically combining the best classifiers for each level, we achieve the best results reported for this corpus.
\end{abstract}

\keywords{Dialog Act Recognition, Hierarchical Classification, Multi-Label Classification, DIHANA Corpus.}

%
%

%
%

\begin{acronym}
    \acro{AL}{Active Learning}
    \acro{AMI}{AMI Meeting Corpus}
    \acro{ASR}{Automatic Speech Recognition}
    \acro{CHS}{CallHome Spanish}
    \acro{CNN}{Convolutional Neural Network}
    \acro{DNN}{Deep Neural Network}
    \acro{DRLM}{Discourse Relation Language Model}
    \acro{DSTC4}{Dialog State Tracking Challenge 4}
    \acro{FCT}{Funda\c{c}\~{a}o para a Ci\^{e}ncia e a Tecnologia}
    \acro{GloVe}{Global Vectors for Word Representation}
    \acro{GPU}{Graphics Processing Unit}
    \acro{GRU}{Gated Recurrent Unit}
    \acro{HCNN}{Hierarchical Convolutional Neural Network}
    \acro{HMM}{Hidden Markov Model}
    \acro{k-NN}{k-Nearest Neighbors}
    \acro{LSA}{Latent Semantic Analysis}
    \acro{LSTM}{Long Short-Term Memory}
    \acro{LVRNN}{Latent Variable Recurrent Neural Network}    
    \acro{MapTask}{HCRC Map Task Corpus}
    \acro{MRDA}{ICSI Meeting Recorder Dialog Act Corpus}
    \acro{NLU}{Natural Language Understanding}
    \acro{NLP}{Natural Language Processing}
    \acro{NN}{Neural Network}
    \acro{POS}{Part of Speech}
    \acro{QA}{Question Answering}
    \acro{RNN}{Recurrent Neural Network}
    \acro{RNNLM}{Recurrent Neural Network Language Model}
    \acro{SVM}{Support Vector Machine}
    \acro{SwDA}{Switchboard Dialog Act Corpus}
    \acro{WoZ}{Wizard of Oz}
\end{acronym}

%
%

%
%

\section{Introduction}
\label{sec:introduction}

It is valuable for a dialog system to identify the intention behind its conversational partners' words since it provides an important cue concerning the information contained in a segment and how it should be interpreted. According to \citet{Searle1969}, that intention is revealed by dialog acts, which are the minimal units of linguistic communication. Consequently, automatic dialog act recognition is an important task in the context of \ac{NLU}, which has been widely explored over the years on multiple corpora with different characteristics. Still, recently, most studies have focused on English data and, more specifically, on the \ac{SwDA}~\citep{Jurafsky1997}, since it is the largest annotated corpus and its label set is independent from both task and domain. However, there are corpora and annotation schemes that pose problems in the context of dialog act recognition which are not covered by the \ac{SwDA} corpus and its SWBD-DAMSL annotations. With this in mind, in this article we explore the DIHANA corpus~\citep{Benedi2006}, which features interactions in Spanish between humans and a \ac{WoZ} dialog system. In the context of dialog act recognition, the differentiating aspect of this corpus is its three-level annotation scheme, in which the top level refers to the generic task-independent dialog act and the others complement it with task-specific information. Additionally, while each segment has a single top-level label, it may have none or multiple labels on the other levels. Thus, the DIHANA corpus allows us to approach dialog act recognition as both a hierarchical and a multi-label classification problem.

Similarly to other text classification tasks, such as news categorization and sentiment analysis~\citep{Kim2014,Conneau2017}, most of the recent approaches on dialog act recognition take advantage of \acp{DNN}. We provide an overview of these approaches in Section~\ref{ssec:sota}. However, overall, they use a \ac{RNN}- or \ac{CNN}-based approach to generate a representation of the segment from the embedding representation of its words and then use the information present in that representation to obtain the classification of the segment. The distinction between \ac{RNN}- and \ac{CNN}-based approaches is relevant since they are able to capture different information. The first focus on identifying relevant word sequences, including long range dependencies. On the other hand, the latter focus on identifying relevant word patterns by inspecting limited context windows surrounding each word. Additionally, the top performing approaches on dialog act recognition do not consider each segment on its own, but rather in combination with context information from both the surrounding segments and concerning the speakers.

Considering the characteristics of the DIHANA corpus and the state-of-the art approaches on single-label dialog act recognition, in this article we explore different aspects. First, we assess whether those approaches perform similarly on a language other than English, by using them to predict the task-independent labels of the top level. Then, we explore their applicability in the multi-label classification scenarios of the other levels. Furthermore, since those levels refer to different task-specific aspects, we also assess how context information from the preceding segments influences the ability to predict each of those aspects. Similarly, we assess how that ability is influenced by information from the upper levels in the hierarchy. Finally, we explore the hierarchical combination of the best approaches for each level and compare its performance with that of the flat approach that was used on previous studies on the corpus.

In the remainder of the article we start by providing an overview of related work in Section~\ref{sec:related}. First, we provide an overview on existing corpora for dialog act recognition in Section~\ref{ssec:corpora}. Then, we discuss the state-of-the-art approaches on dialog act recognition in Section~\ref{ssec:sota}. Additionally, previous studies on dialog act recognition on Spanish data are summarized in Section~\ref{ssec:spanish}. Then, in Section~\ref{sec:setup}, we describe our experimental setup. We start by describing the DIHANA corpus and its dialog act annotations in Section~\ref{ssec:dihana}. Section~\ref{ssec:arch} presents the generic network architecture used in our experiments and describes what changes between experiments. Finally, Section~\ref{ssec:eval} introduces the training and evaluation procedures according to the level of the hierarchy in focus. The results achieved by our experiments on each of those levels, as well as their combination, are presented and discussed in Section~\ref{sec:results}. Finally, Section~\ref{sec:conclusions} states the most important conclusions that can be drawn from the experiments described in this article and provides pointers for future work. 

%
%

%
%

\section{Related Work}
\label{sec:related}

As previously stated, automatic dialog act recognition is a task that has been widely explored over the years on multiple corpora with different characteristics and using a variety of classical machine learning approaches, from \acp{HMM}~\citep{Stolcke2000} to \acp{SVM}~\citep{Gamback2011}. The article by \citet{Kral2010} provides an overview of most of those approaches on the task. However, recently, most approaches take advantage of \ac{DNN} architectures. Below, we present an overview of such approaches. Additionally, since our study focuses on the DIHANA corpus~\citep{Benedi2006}, we also present previous approaches on dialog act recognition on Spanish data. However, first, we provide an overview on existing corpora for dialog act recognition.

\subsection{Corpora for Dialog Act Recognition}
\label{ssec:corpora}

Multiple corpora have been annotated in terms of dialog acts. Table~\ref{tab:corpora} presents a non-exhaustive set of those corpora and their characteristics. We can see that multiple domains, languages, and kinds of interaction are covered, which enables the assessment of the generalization capabilities of dialog act recognition approaches in multiple scenarios. However, on the other hand, the used tag sets are not standardized among corpora. In fact, there are even different tag sets for the same corpus. This means that the sets were developed with different objectives and have different hierarchies and levels of abstraction, which makes cross-corpora and generalization experiments hard to perform. This is particularly problematic when the used tag sets are domain-dependent, since they cannot be applied to corpora in other domains.

\begin{table*}[ht]
\begin{center}
    \resizebox{\textwidth}{!}{
    \begin{tabular}{l l l l r r c}
        \toprule
        \textbf{Corpus}                    & \textbf{Interaction} & \textbf{Domain} & \textbf{Language} & \textbf{Segments} & \textbf{Tags} & \textbf{DD} \tabularnewline
        \midrule
        \acs{SwDA}~\citep{Jurafsky1997}    & Human                & Open            & English           &              220k &       41 - 44 &           N \tabularnewline
        \acs{MRDA}~\citep{Shriberg2004}    & Human                & Meetings        & English           &              106k &   5 / 11 + 39 &           N \tabularnewline
        \acs{AMI}~\citep{Carletta2005}     & Human                & Meetings        & English           &              102k &            15 &           N \tabularnewline
        VERBMOBIL~\citep{Kay1992}          & Human                & Schedules       & Multiple          &               59k &       42 / 33 &           M \tabularnewline
        \acs{CHS}~\citep{Levin1998}        & Human                & Open            & Spanish           &               45k &       10 / 37 &           N \tabularnewline
        \acs{DSTC4}~\citep{Kim2016}        & Human                & Travel          & English           &               31k &            89 &           Y \tabularnewline
        \acs{MapTask}~\citep{Anderson1991} & Human                & Routes          & English           &               27k &            12 &           N \tabularnewline
        DIHANA~\citep{Benedi2006}          & \ac{WoZ}             & Trains          & Spanish           &               23k &  11 + 10 + 13 &           M \tabularnewline
        LEGO~\citep{Schmitt2012}           & Machine              & Buses           & English           &               14k &       22 + 28 &           Y \tabularnewline
        NESPOLE~\citep{Costantini2002}     & Human                & Travel          & Multiple          &                8k &       67 + 91 &           M \tabularnewline
        DIME~\citep{Villasenor2001}        & \ac{WoZ}             & Kitchen Design  & Spanish           &                5k &       15 + 15 &           M \tabularnewline
        \bottomrule
    \end{tabular}
    }
\end{center}
\caption{Corpora annotated with dialog act information, ordered by number of segments. The values for the number of segments are rounded. The interaction column states whether the dialogs are between humans or if there is a dialog system involved. In the latter case it distinguishes between \ac{WoZ} scenarios and interactions with a real machine. In the tags column, the / and - symbols refer to alternative tag sets, while the + symbol refers to different levels of annotation. The last column, DD, states whether the tag set is domain-dependent (Y), domain-independent (N), or mixed (M).}
\label{tab:corpora}
\end{table*}

Concerning alternative tag sets for the same corpus, while those of \ac{SwDA}, the \ac{MRDA}, and \ac{CHS} are just compressed versions of the original sets, the two tag sets used to annotate VERBMOBIL are disjoint. Furthermore, the first one includes domain-dependent labels~\citep{Jekat1995}, while the second is completely domain-independent~\citep{Alexandersson1998}.

Multiple corpora have complementary tag sets which refer to different aspects. For instance, \ac{MRDA}, DIHANA, and NESPOLE have a set of generic labels which can be specialized using labels from different sets. However, while in the first case the specialized labels are still domain-independent, in the remaining two the generic labels are complemented with domain-specific information at different levels. On the DIME corpus, the two tag sets refer to different aspects of the dialog, namely, obligations and grounding. Finally, the LEGO corpus has independent tag sets for user and system segments. 

In an attempt to standardize dialog act annotation and, consequently, set the ground for more comparable research in the area, \citet{Bunt2012} defined the ISO 24617-2 standard. According to it, dialog act annotations should be performed on functional segments rather than on turns or utterances~\citep{Carroll1978}. Furthermore, the annotation of each segment does not consist of a single label, but rather of a complex structure containing information about the participants, relations with other functional segments, the semantic dimension of the dialog act, its communicative function, and optional qualifiers concerning certainty, conditionality, partiality, and sentiment. However, annotating all of these aspects is an exhaustive process and, consequently, the amount of data annotated according to the standard is still reduced and, in many cases, not all of the aspects are considered~\citep{Petukhova2014,Bunt2016,Ribeiro2016}.

As previously stated, most recent studies on automatic dialog act recognition take advantage of different \ac{DNN} architectures. Such approaches require large amounts of data to train. Consequently, the automatic prediction of dialog acts as defined by the standard has only been approached in a few studies~\citep{Ribeiro2015,Mezza2018}. On the other hand, \ac{SwDA} is the most explored corpus for the task, since it is the one with the highest number of annotated segments, it features open-domain dialogs, and its tag set is domain-independent. Thus, the conclusions drawn from experiments on it are expected to generalize well to other scenarios.

\subsection{State-of-the-Art on Dialog Act Recognition}
\label{ssec:sota}

The approaches that achieve highest performance on the dialog act recognition task are based on \acp{DNN}. Thus, in this section, we focus on studies that use such approaches. To our knowledge, the first of those studies was that by \citet{Kalchbrenner2013}. The described approach uses a \ac{CNN}-based approach to generate segment representations from randomly initialized word embeddings. Then, it uses a \ac{RNN}-based discourse model that combines the sequence of segment representations with speaker information and outputs the corresponding sequence of dialog acts. By limiting the discourse model to consider information from the two preceding segments only, this approach achieved 73.9\% accuracy on the \ac{SwDA} corpus. 

\citet{Lee2016} compared the performance of a \ac{LSTM} unit against that of a \ac{CNN} to generate segment representations from pre-trained embeddings of its words. In order to generate the corresponding dialog act classifications, the segment representations were then fed to a 2-layer feed-forward network, in which the first layer normalizes the representations and the second selects the class with highest probability. In their experiments, the \ac{CNN}-based approach consistently led to similar or better results than the \ac{LSTM}-based one. The architecture was also used to provide context information from up to two preceding segments at two levels. The first level refers to the concatenation of the representations of the preceding segments with that of the current segment before providing it to the feed-forward network. The second refers to the concatenation of the normalized representations before providing them to the output layer. This approach achieved 65.8\% accuracy on the \ac{DSTC4} corpus, 84.6\% on \ac{MRDA} with 5 classes~\citep{Ang2005}, and 71.4\% on \ac{SwDA}. However, the influence of context information varied across corpora.

\citet{Ji2016} explored the combination of positive aspects of \ac{NN} architectures and probabilistic graphical models. They used a \ac{DRLM} that combines a \ac{RNNLM}~\citep{Mikolov2010} to model the sequence of words in the dialog with a latent variable model over shallow discourse structure to model the relations between adjacent segments which, in this context, represent the dialog acts. This way, the model can perform word prediction using discriminatively-trained vector representations while maintaining a probabilistic representation of a targeted linguistic element, such as the dialog act. In order to function as a dialog act classifier, the model was trained to maximize the conditional probability of a sequence of dialog acts given a sequence of segments, achieving 77.0\% accuracy on \ac{SwDA}.

\citet{Tran2017a} used a hierarchical \ac{RNN} with an attentional mechanism to predict the dialog act classifications of a whole dialog. The model is hierarchical, since it includes an utterance-level \ac{RNN} to generate the representation of the utterance from its tokens and another to generate the sequence of dialog act labels from the sequence of utterance representations. The attentional mechanism is between the two, since it uses information from the dialog-level \ac{RNN} to identify the most important tokens in the current utterance and filter its representation. Using this approach they achieved 74.5\% accuracy on \ac{SwDA} and 63.3\% on the \ac{MapTask} corpus. Later, they were able to improve the performance on \ac{SwDA} to 75.6\% by propagating uncertainty information concerning the previous predictions~\citep{Tran2017c}. Additionally, they experimented with gated attention in the context of a generative model, achieving 74.2\% on \ac{SwDA} and 65.94\% on \ac{MapTask}~\citep{Tran2017b}.

The previous studies explored the use of a single recurrent or convolutional layer to generate the segment representation from those of its words. However, the approaches with highest performance on the task use multiple of those layers. On the one hand, \citet{Khanpour2016} achieved their best results using a segment representation generated by concatenating the outputs of a stack of 10 \ac{LSTM} units at the last time step. This way, the model is able to capture long distance relations between tokens. On the other hand, \citet{Liu2017} generated the segment representation by combining the outputs of three parallel \acp{CNN} with different context window sizes, in order to capture different functional patterns. In both cases, pre-trained word embeddings were used as input to the network. Overall, from the reported results, it is not possible to state which is the top performing segment representation approach since the evaluation was performed on different subsets of \ac{SwDA}. Still, \citet{Khanpour2016} reported 73.9\% accuracy on the validation set and 80.1\% on the test set, while \citet{Liu2017} reported 74.5\% and 76.9\% accuracy on the two sets used to evaluate their experiments. Additionally, \citet{Khanpour2016} reported 86.8\% accuracy on \ac{MRDA}.

Additionally, \citet{Liu2017} explored the use of context information concerning speaker changes and from the surrounding segments. The first was provided as a flag and concatenated to the segment representation. Concerning the latter, they explored the use of discourse models, as well as of approaches that concatenated the context information directly to the segment representation. The discourse models transform the model into a hierarchical one by generating a sequence of dialog act classifications from the sequence of segment representations. Thus, when predicting the classification of a segment, the surrounding ones are also taken into account. However, when the discourse model is based on a \ac{CNN} or a bidirectional \ac{LSTM} unit, it considers information from future segments, which is not available to a dialog system. Still, even when relying on future information, the approaches based on discourse models performed worse than those that concatenated the context information directly to the segment representation. In this sense, providing that information in the form of the classification of the surrounding segments led to better results than using their words, even when those classifications were obtained automatically. This conclusion is in line with what we had shown in our previous study using \acp{SVM}~\citep{Ribeiro2015}. Furthermore, both studies have shown that, as expected, the first preceding segment is the most important and that the influence decays with the distance. Using the setup with gold standard labels from three preceding segments, the results on the two sets used to evaluate the approach improved to 79.6\% and 81.8\%, respectively.

It is important to make some remarks concerning tokenization and token representation. In all the previously described studies, tokenization was performed at the word level. Furthermore, with the exception of the first study~\citet{Kalchbrenner2013}, which used randomly initialized embeddings, and those by \citet{Tran2017b,Tran2017a,Tran2017c}, for which the embedding approach was not disclosed, the representation of those words was given by pre-trained embeddings. \citet{Khanpour2016} compared the performance when using Word2Vec~\citep{Mikolov2013} and \ac{GloVe}~\citep{Pennington2014} embeddings trained on multiple corpora. Although both embedding approaches capture information concerning words that commonly appear together, the best results were achieved using Word2Vec embeddings. In terms of dimensionality, \citet{Khanpour2016} achieved the best results when using 150-dimensional embeddings. However, 200-dimensional embeddings were used in other studies~\citep{Lee2016,Liu2017}, which was not one of the compared values.

The approaches described in all of the previous studies perform tokenization at the word level. However, we have shown that there are also important cues for intention at a sub-word level which can only be captured when using a finer-grained tokenization, such as at the character-level~\citep{Ribeiro2018}. The cues at that level mostly refer to aspects concerning the morphology of words, such as lemmas and affixes. To capture that information, we adapted the \ac{CNN}-based segment representation approach by \citet{Liu2017} to use characters instead of words as tokens. This way, we were able to explore context windows of different sizes to capture those different morphological aspects. In this sense, our best results were achieved when using three parallel \acp{CNN} with window sizes 3, 5, and 7, which are able to capture affixes, lemmas, and inter-word relations, respectively. Using this approach we achieved 76.8\% and 73.2\% accuracy on the validation and test sets of \ac{SwDA}, respectively. These results are in line with those of the word-level approach. However, the combination of the two levels improved the results to 78.0\% and 74.0\%, respectively, which shows that character- and word-level tokenizations provide complementary information. Finally, by including context information from three preceding segments, we improved the results to 82.0\% accuracy on the validation set and 79.0\% on the test set.

\subsection{Dialog Act Recognition on Spanish Data}
\label{ssec:spanish}

Research on dialog act recognition on Spanish data has been mainly performed on two corpora {---} DIHANA and \ac{CHS}. Both feature spontaneous telephonic dialogs. However, as shown in Table~\ref{tab:corpora}, while the dialogs from the first are between humans and a \ac{WoZ} dialog system, the ones from the latter are between humans. Furthermore, while \ac{CHS} is annotated using task-independent labels, DIHANA is annotated using a three-level hierarchical label scheme, in which the first level refers to the generic task-independent dialog act and the others complement it with task-specific information. There is also a series of experiments on dialog act recognition on the DIME corpus~\citep{Coria2005,Coria2006,Coria2009}. However, these focused on using prosodic information to predict specific subsets of the obligations and grounding dialog acts that the corpus is annotated with. Since our work focuses on dialog act recognition from textual data, we will only provide further detail on the studies performed on the first two corpora. 

The first dialog act recognition experiments on the DIHANA corpus employed \acp{HMM} using both prosodic~\citep{Tamarit2008} {---} energy and pitch {---} and textual~\citep{Martinez-Hinarejos2008} {---} n-grams {---} features. The first achieved 60.70\% accuracy on the first level, while the latter achieved 93.40\% on the combination of the first two levels and 89.70\% on the combination of all levels. The latter study, as well as a more recent one~\citep{Martinez-Hinarejos2015}, also explored the recognition of dialog acts on unsegmented turns using n-gram transducers. However, in those cases, the focus was on the segmentation process and the classification approaches did not differ from the previous. Finally, the approach which obtained best results on the manually segmented dialogs was based on \acp{SVM} using n-grams, the presence of wh-words, and punctuation, as well as context information from three preceding segments as features~\citep{Gamback2011}. This approach also applied \ac{AL} to reduce the amount of data required for training, achieving 94.08\% accuracy on the combination of the first two levels and 90.97\% on the combination of all levels.

Similarly to the DIHANA corpus, the first dialog act recognition experiments on the \ac{CHS} corpus also employed \acp{HMM} with different types of n-gram~\citep{Levin1999,Ries1999}. The latter study improved the results by combining the \acp{HMM} with \acp{NN} using unigrams and \ac{POS} tags as features, achieving 76.1\% accuracy. The task was also approached using \ac{LSA} in three different studies~\citep{Serafin2003,Serafin2004,DiEugenio2010}. The first used both plain \ac{LSA} and multiple adaptations based on clustering and the incorporation of features concerning the preceding dialog acts. However, there was no improvement over plain \ac{LSA}, which achieved 65.36\% accuracy on the tag set with 37 classes and 68.91\% on the compressed set of 10 classes. On the other hand, the remaining studies experimented with multiple syntactic and dialog related features and were able to improve the results of plain \ac{LSA}, up to 77.74\% and 81.27\%, respectively. In the last study, these results were further improved to 80.34\% and 82.88\% by applying an instance-based learning approach, namely \ac{k-NN}, to the reduced semantic spaces computed by \ac{LSA}. However, in both cases, the improvements were achieved using features concerning the dialog game, that is, the generic intention of the whole dialog, and whether the speaker is taking initiative or replying or providing feedback to the other speaker. Although in general the dialog game is known, there are also cases in which a dialog system is not aware of it. Furthermore, identifying whether a speaker is taking initiative, replying, or providing feedback can be seen as a simplification of the dialog act recognition task. Thus, it is not fair to use that information if it is not obtained automatically as well. Finally, the corpus was also explored in domain adaptation experiments for dialog act classification using a reduced set of classes~\citep{Margolis2010}.

%
%

%
%

\section{Experimental Setup}
\label{sec:setup}

We want to assess whether the top performing approaches described in the previous section perform similarly on a language other than English. Furthermore, we want to explore their applicability in the multi-label classification scenarios posed by the two bottom levels of the DIHANA corpus dialog act annotations. Since those levels refer to different task-specific aspects, we also assess how context information from the preceding segments influences the ability to predict each of those aspects. Similarly, we assess how that ability is influenced by information from the upper levels. Finally, we want to assess whether the hierarchical combination of the best approaches for each level is able to outperform the flat approach that was used in previous studies on the corpus.

In this section we describe our experimental setup, starting with a description of the DIHANA corpus and its dialog act annotations. Then, we present the generic architecture used in our experiments and explain how it changes according to the aspect and the characteristics of the level in focus, especially considering the differences between single- and multi-label classification. Finally, we describe our training and evaluation approaches, including the differences in the metrics used for single- and multi-label problems.

%
%

\subsection{Dataset}
\label{ssec:dihana}

The DIHANA corpus~\citep{Benedi2006} consists of 900 dialogs between 225 human speakers and a \ac{WoZ} telephonic train information system. There are 6,280 user turns and 9,133 system turns, with a vocabulary size of 823 words and a total of 48,243 words. The turns were manually transcribed, segmented, and annotated with dialog acts~\citep{Alcacer2005}. The total number of annotated segments is 23,547, with 9,715 corresponding to user segments and 13,832 to system segments. One of the annotated dialogs is shown in Figure~\ref{fig:dialog}.

\begin{figure*}[ht]
    \begin{framed}
    \begin{dialogue}
        \speak{System} Bienvenido al servicio de informacion de trenes ¿En qué puedo ayudarle? \\
            \textit{(Welcome to the train information system. How can I help you?)} \\
            L1: \textit{Apertura}, L2: \textit{Nil}, L3: \textit{Nil}
		\speak{User} Quería ir a La Coruña \textit{(I would like to go to Corunna)} \\
            L1: \textit{Pregunta}, L2: \textit{Hora Salida}, L3: \textit{Destino}
        \speak{User} y quería saber los horarios que hay de los trenes para el lunes. \\
            \textit{(and I would like to know the schedule for the trains on Monday.)} \\
            L1: \textit{Pregunta}, L2: \textit{Hora Salida}, L3: \textit{Día}
        \speak{System} ¿Quiere viajar desde desde La Coruña? \textit{(Do you want to depart from Corunna?)} \\
            L1: \textit{Confirmación}, L2: \textit{Origen}, L3: \textit{Origen}
        \speak{User} No. \textit{(No.)} \\
            L1: \textit{Negación}, L2: \textit{Origen}, L3: \textit{Nil}
        \speak{User} Desde Bilbao. \textit{(From Bilbao.)} \\
            L1: \textit{Respuesta}, L2: \textit{Hora Salida}, L3: \textit{Origen}
        \speak{System} Quiere horarios de trenes a La Coruña \textit{(You want train schedules to Corunna)} \\
            L1: \textit{Confirmación}, L2: \textit{Hora Salida}, L3: \textit{Destino}
        \speak{System} ¿desde Bilbao? \textit{(from Bilbao?)} \\
            L1: \textit{Confirmación}, L2: \textit{Origen}, L3: \textit{Origen}     
        \speak{User} Sí. \textit{(Yes.)} \\
            L1: \textit{Afirmación}, L2: \textit{Origen}, L3: \textit{Nil}
        \speak{System} Le consulto horarios de trenes con salida en Bilbao destino La Coruña saliendo el lunes, día 16 de febrero de 2004. \\
            \textit{(I'm looking for trains departing from Bilbao to Corunna on Monday, February 16, 2004.)} \\
            L1: \textit{Confirmación}, L2: \textit{Hora Salida}, L3: \textit{Destino, Día, Origen}
        \speak{System} Un momento por favor. \textit{(A moment please.)} \\
            L1: \textit{Espera}, L2: \textit{Nil}, L3: \textit{Nil}
        \speak{System} El único tren que realiza el trayecto es un Diurno que sale a las 9 y 25 de la mañana. \\
            \textit{(The only train that makes that journey is a Diurno that departs at 9:25 a.m.)} \\
            L1: \textit{Respuesta}, L2: \textit{Hora Salida}, L3: \textit{Hora Salida, Número Trenes, Tipo Tren}
        \speak{System} ¿Desea algo más? \textit{(Do you need anything else?)} \\
            L1: \textit{Nueva Consulta}, L2: \textit{Nil}, L3: \textit{Nil}
        \speak{User} Sí, ¿me puede decir el precio del billete? \textit{(Yes. Can you tell me the cost of the ticket?)} \\
            L1: \textit{Pregunta}, L2: \textit{Precio}, L3: \textit{Nil}
		\speak{System} El precio de ese tren en clase turista es 36 euros el billete de ida. Recuerde que el billete de ida y vuelta siempre supone un ahorro del 20 por ciento. \\
            \textit{(The price of that train is 36 euros one-way in economy class. Remember that return tickets have a 20 percent discount.)} \\
            L1: \textit{Respuesta}, L2: \textit{Precio}, L3: \textit{Clase Billete, Precio}
		\speak{System} ¿Desea algo más? \textit{(Do you need anything else?)} \\
            L1: \textit{Nueva Consulta}, L2: \textit{Nil}, L3: \textit{Nil}
        \speak{User} No. Creo que no. \textit{(No, I don't think so.)} \\
            L1: \textit{Cierre}, L2: \textit{Nil}, L3: \textit{Nil}
		\speak{System} Gracias por utilizar este servicio. Feliz viaje. \\
            \textit{(Thank you for using this service. Have a good journey.)} \\
            L1: \textit{Cierre}, L2: \textit{Nil}, L3: \textit{Nil}
    \end{dialogue}
    \end{framed}
    \caption{A DIHANA corpus dialog. Each segment is followed by its English translation in brackets and its three-level dialog act annotation.}
    \label{fig:dialog}
\end{figure*}

The dialog act annotations are hierarchically decomposed in three levels~\citep{Martinez-Hinarejos2002}. The first level (L1) represents the generic intention of the segment, independently of task details, while the remaining (L2 and L3) represent task-specific information. The first level has 11 labels, distributed according to Table~\ref{tab:l1distr}. In that table we can see that two of the labels are exclusive to user segments {---} \textit{Acceptance} and \textit{Rejection} {---} and four to system segments {---} \textit{Opening}, \textit{Waiting}, \textit{New Consult}, and \textit{Confirmation}. Furthermore, the most common label, \textit{Question}, covers 27\% of the segments.   

\begin{table}[ht]
\begin{center}
    \resizebox{\columnwidth}{!}{
    \begin{tabular}{l r r r r}
        \toprule
        \textbf{Label} & \textbf{User} & \textbf{System} & \textbf{Total} & \textbf{\%} \tabularnewline
        \midrule
        Pregunta \textit{\small (Question)}           & 5,474 &     864 &  6,338 & 27 \tabularnewline
        Respuesta \textit{\small (Answer)}            & 1,839 &   2,446 &  4,285 & 18 \tabularnewline
        Confirmación \textit{\small (Confirmation)}   &     0 &   3,629 &  3,629 & 15 \tabularnewline
        Nueva Consulta \textit{\small (New Consult)}  &     0 &   2,474 &  2,474 & 11 \tabularnewline
        Espera \textit{\small (Waiting)}              &     0 &   1,948 &  1,948 &  8 \tabularnewline
        Cierre \textit{\small (Closing)}              &   927 &     900 &  1,827 &  8 \tabularnewline
        Afirmación \textit{\small (Acceptance)}       &   990 &       0 &    990 &  4 \tabularnewline
        Apertura \textit{\small (Opening)}            &     0 &     900 &    900 &  4 \tabularnewline
        No Entendido \textit{\small (Not Understood)} &     4 &     653 &    657 &  3 \tabularnewline
        Negación \textit{\small (Rejection)}          &   340 &       0 &    340 &  1 \tabularnewline
        Indefinida \textit{\small (Undefined)}        &   141 &      18 &    159 &  1 \tabularnewline
        \bottomrule
    \end{tabular}
    }
\end{center}
\caption{Level 1 label distribution. The English translation of each label is in brackets.}
\label{tab:l1distr}
\end{table}

Although they share most labels, the two task-specific levels of the hierarchy focus on different information. While the second level is related to the kind of information that is implicitly focused in the segment, the third level is related to the kind of information that is explicitly referred to in the segment. For instance, consider the segment \textit{''I’m looking for trains departing from Bilbao to Corunna on Monday, February 16, 2004.''} extracted from the dialog in Figure~\ref{fig:dialog}. Since it reveals the intention of finding a train schedule, it has \textit{Departure Time} as a Level 2 label. However, since that departure time is not explicitly refered in the segment, that label is not part of its Level 3 labels. On the other hand, the segment explicitly refers a departure place, a destination, and a date. Thus, it has the corresponding Level 3 labels {---} \textit{Origin}, \textit{Destination}, and \textit{Day}.

The label distributions in both levels are shown in Table~\ref{tab:l23distr}. We can see that there are 10 common labels and three additional ones on Level 3 {---} \textit{Order Number}, \textit{Number of Trains}, and \textit{Trip Type}. Furthermore, both levels have the \textit{Nil} label, which represents the absence of label in that level. In this sense, we can see that only 63\% of the segments have Level 2 labels, and that the percentage is even lower, 52\%, when considering Level 3 labels. This is mainly due to the fact that segments labeled as \textit{Opening}, \textit{Closing}, \textit{Undefined}, \textit{Not Understood}, \textit{Waiting}, and \textit{New Consult} on the first level cannot have labels on the remaining levels. Finally, it is important to refer that while each segment may only have one Level 1 label, it may have multiple Level 2 and Level 3 labels.

\begin{table*}[ht]
\begin{center}
    \resizebox{\textwidth}{!}{
    \begin{tabular}{l r r r r | l r r r r}
        \toprule
        \multicolumn{5}{c|}{\textbf{Level 2}}                                                 & \multicolumn{5}{c}{\textbf{Level 3}}                                                      \tabularnewline
        \textbf{Label} & \textbf{User} & \textbf{System} & \textbf{Total} & \textbf{\%} & \textbf{Label} & \textbf{User} & \textbf{System} & \textbf{Total} & \textbf{\%} \tabularnewline
        \midrule
        Nulo \textit{\small (Nil)}                   & 1,923 &  6,893 &  8,816 &  37 & Nulo \textit{\small (Nil)}                       & 2,954 &  8,317 & 11,271 &  48 \tabularnewline
        Hora Salida \textit{\small (Departure Time)} & 3,309 &  3,523 &  7,432 &  32 & Destino \textit{\small (Destination)}            & 1,631 &  2,079 &  3,710 &  16 \tabularnewline
        Precio \textit{\small (Fare)}                & 2,071 &  1,267 &  3,338 &  14 & Día \textit{\small (Day)}                        & 1,881 &  1,778 &  3,659 &  16 \tabularnewline
        Día \textit{\small (Day)}                    & 1,026 &    923 &  1,949 &   8 & Origen \textit{\small (Origin)}                  &   896 &  2,085 &  2,981 &  13 \tabularnewline
        Origen \textit{\small (Origin)}              &   477 &    480 &    957 &   4 & Hora Salida \textit{\small (Departure Time)}     &   692 &  1,633 &  2,325 &  10 \tabularnewline
        Destino \textit{\small (Destination)}        &   452 &    400 &    852 &   4 & Número Trenes \textit{\small (Number of Trains)} &     0 &  1,863 &  1,863 &   8 \tabularnewline
        Tipo Tren \textit{\small (Train Type)}       &   317 &    226 &    543 &   2 & Tipo Tren \textit{\small (Train Type)}           &   544 &  1,253 &  1,797 &   8 \tabularnewline
        Hora Llegada \textit{\small (Arrival Time)}  &    90 &     88 &    178 &   1 & Número Orden \textit{\small (Order Number)}      &    84 &    950 &  1,034 &   4 \tabularnewline
        Tiempo Recorrido \textit{\small (Duration)}  &    14 &     15 &     29 & 0.1 & Clase Billete \textit{\small (Ticket Class)}     &   129 &    766 &    895 &   4 \tabularnewline
        Clase Billete \textit{\small (Ticket Class)} &    15 &     12 &     27 & 0.1 & Precio \textit{\small (Fare)}                    &    47 &    731 &    778 &   3 \tabularnewline
        Servicio \textit{\small (Service)}           &     3 &      5 &      8 &   0 & Hora Llegada \textit{\small (Arrival Time)}      &   199 &    490 &    689 &   3 \tabularnewline
                                                     &       &        &        &     & Tipo Viaje \textit{\small (Trip Type)}           &   643 &      0 &    643 &   3 \tabularnewline 
                                                     &       &        &        &     & Servicio \textit{\small (Service)}               &    15 &      4 &     19 & 0.1 \tabularnewline
                                                     &       &        &        &     & Tiempo Recorrido \textit{\small (Duration)}      &     0 &     14 &     14 & 0.1 \tabularnewline
        \bottomrule
    \end{tabular}
    }
\end{center}
\caption{Level 2 and Level 3 label distributions. The English translation of each label is in brackets.}
\label{tab:l23distr}
\end{table*}

As a final remark, it is important to refer that some Level 2 {---} \textit{Duration}, \textit{Ticket Class}, and \textit{Service} {---} and Level 3 {---} \textit{Service} and \textit{Duration} {---} labels only occur in 0.1\% of the segments or less. Thus, these are hard to predict using machine learning approaches that focus on maximizing the overall accuracy on the corpus. 

%
%

%
%

\subsection{Network Architecture}
\label{ssec:arch}

\begin{figure}[ht]
	\centering
	\includegraphics[width=\columnwidth]{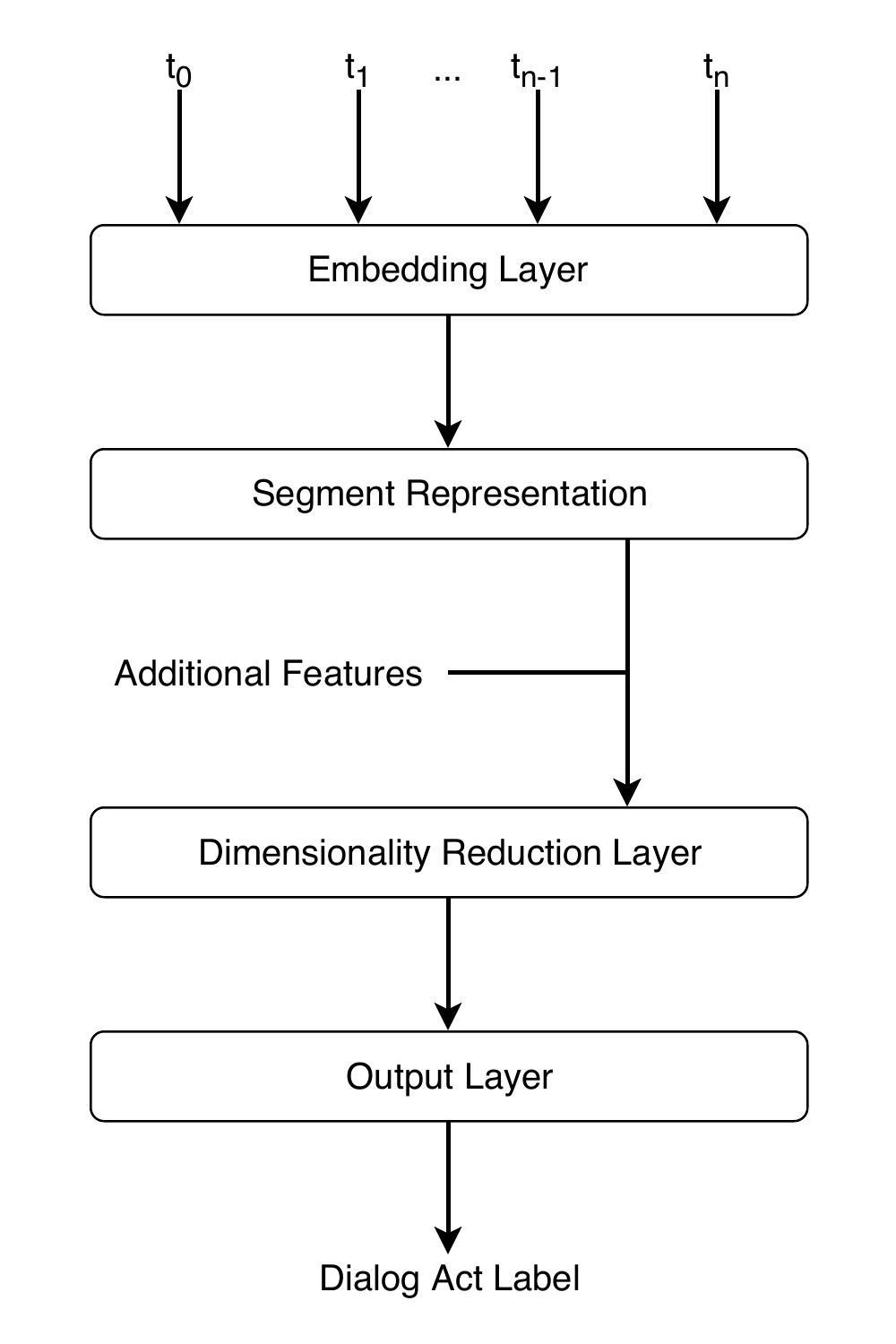}
	\caption{The generic architecture of the networks used in our experiments. $t_i$ corresponds to the $i$-th token.}
	\label{fig:arch}
\end{figure}

Since we want to assess the performance of different \ac{DNN}-based approaches on dialog act recognition on the DIHANA corpus, we must define a common ground for comparison. Thus, we use a generic network architecture, shown in Figure~\ref{fig:arch}, which is based on those of the top performing approaches referred to in Section~\ref{ssec:sota}. The generic approach to obtain a dialog act classification for a segment is as follows: First, the segment is split into tokens, which are passed to an embedding layer. Then, the sequence of token embeddings is passed to the segment representation approach. The obtained representation can then be concatenated with additional information from other sources before being passed to a dimensionality reduction layer. Finally, the obtained reduced representation is passed to the output layer, which generates the dialog act classification. The motivation for each of these steps and their characteristics according to the level of the hierarchy in focus are described below.

\subsubsection{Embedding Layer}

The input of our network is the sequence of tokens in the segment. Similarly to most previous approaches on dialog act recognition, we perform tokenization at the word level. As shown in our previous study~\citep{Ribeiro2018}, the character level is also able to provide important information. However, for simplification, we do not include it in this study. Furthermore, we ignore punctuation, since it may not be available for a dialog system. The tokens are then passed to the embedding layer to be transformed into a vectorial representation corresponding to their position in the embedding space. In our experiments, we use pre-trained word embeddings obtained by applying Word2Vec~\citep{Mikolov2013} on the Spanish Billion Words Corpus~\citep{Cardellino2016}. Although we have explored embedding spaces with different dimensionality, we only report the results obtained using dimensionality 200, as used by \citet{Liu2017}, since it consistently led to better results than the ones explored by \citet{Khanpour2016}.

\subsubsection{Segment Representation}

The segment representation step generates a vectorial representation of the segment through the combination of the representations of its tokens. As stated in Section~\ref{ssec:sota}, the two approaches with higher performance on dialog act recognition on English data vary on this step. While the approach by \citet{Khanpour2016} is based on \acp{RNN}, the one by \citet{Liu2017} is based on \acp{CNN}. Both have their own advantages, as while the first focuses on capturing information from relevant sequences of tokens, the latter focuses on the context surrounding each token and, thus, captures relevant patterns. Since the different levels in the label hierarchy have different characteristics, we use both approaches in our experiments to assess whether one outperforms the other in every situation or the one with best performance varies according to the level.

\begin{figure}[ht]
	\centering
	\includegraphics[width=\columnwidth]{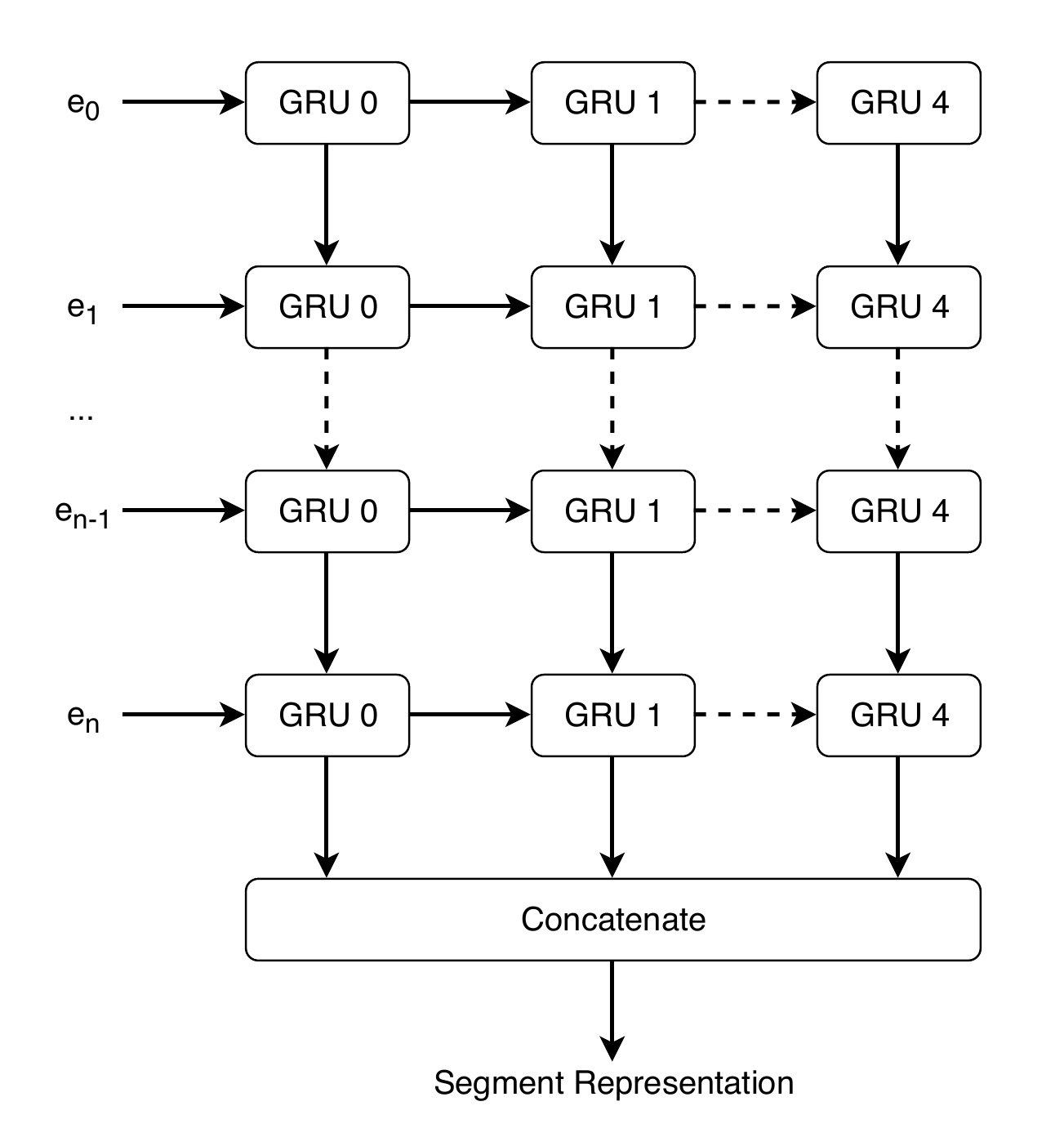}
	\caption{The \ac{RNN}-based segment representation approach. $e_i$ corresponds to the embedding representation of the $i$-th token.}
	\label{fig:rnn}
\end{figure}

As described in Section~\ref{ssec:sota}, the \ac{RNN}-based approach by \citet{Khanpour2016} uses a stack of 10 \ac{LSTM} units. The segment representation is given by the concatenation of the outputs of the 10 \ac{LSTM} units at the last time step, that is, after processing all the tokens in the segment. Using the output at the last time step instead of other pooling operation makes sense, since the recurrent units process the tokens sequentially. Thus, that output contains information from the whole segment. The results reported in this article were obtained using a stack of five \acp{GRU} instead of the stack of 10 \acp{LSTM}, since it led to similar performance with reduced resource consumption on our preliminary experiments. A graphical representation of this approach is shown in Figure~\ref{fig:rnn}.

\begin{figure}[ht]
	\centering
	\includegraphics[width=\columnwidth]{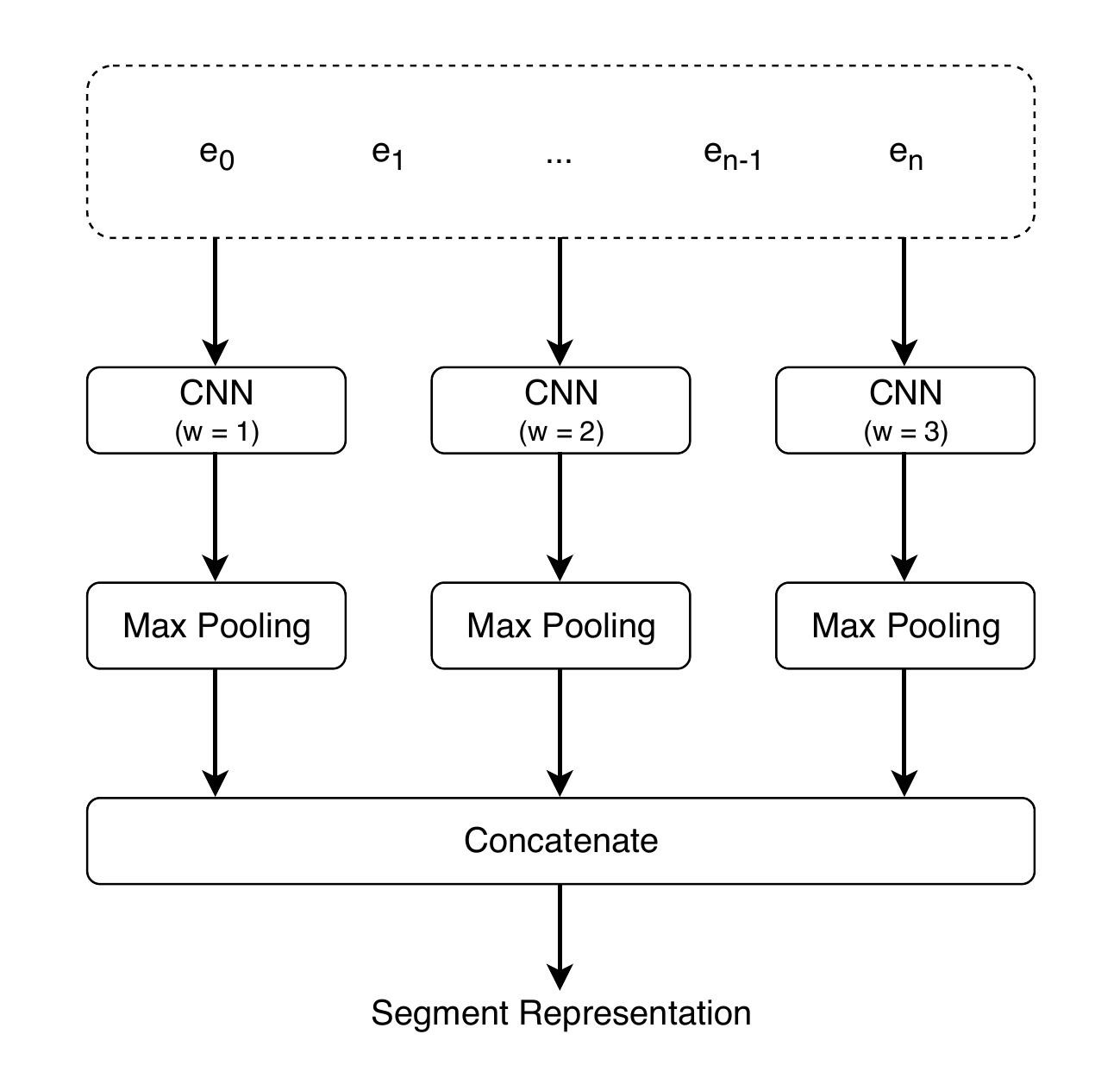}
	\caption{The \ac{CNN}-based segment representation approach. $e_i$ corresponds to the embedding representation of the $i$-th token. The parameter $w$ refers to the size of the context window.}
	\label{fig:cnn}
\end{figure}

Also as described in Section~\ref{ssec:sota}, the \ac{CNN}-based approach by \citet{Liu2017} uses three parallel temporal \acp{CNN} with window sizes between one and three, inclusively. This means that it focuses on sets of at most three consecutive words. A previous study by \citet{Kim2014} used window sizes between three and five, in order to capture relations between more distant words, which were relevant for the approached tasks. Considering the task at hand, the most relevant window sizes depend on the level in focus, as the task-specific dialog acts are typically related to the presence of specific words, while generic dialog acts are more related to the structure of the segment and, consequently, larger windows. Thus, we explore the use of different window sizes for each level. The outputs of the \acp{CNN} suffer a max pooling operation and are afterwards concatenated to generate the segment representation. A graphical representation of the approach is shown in Figure~\ref{fig:cnn}.

\subsubsection{Context Information}
\label{sssec:context}

Previous studies have confirmed the importance of context information provided by the preceding segments for dialog act recognition~\citep{Ribeiro2015,Lee2016,Liu2017}. Additionally, those studies have shown that the influence of preceding segments decays with distance and that the dialog act classification of those segments is more informative than their words. Thus, in our experiments we use the same label-based representation approach used in our study on the \ac{SwDA} corpus~\citep{Ribeiro2015} and also by \citet{Liu2017} to provide context information to the network. That is, the labels from the preceding segments are transformed into the corresponding one-hot encodings and concatenated to the segment representation. Similarly to \citet{Liu2017}, we explore the use of context information from up to three preceding segments, since our previous study has shown that the improvement achieved by using additional segments is negligible. In the context of a dialog system identifying its conversational partner's intention, the system only has access to the preceding segments. Thus, in our experiments we do not use information extracted from future segments. It is important to refer that we use the manual annotations of the segments to provide the context information. Thus, the obtained results represent an upper bound for the approach. We did not use automatic labels in our experiments since both our study and that by \citet{Liu2017} have shown that this approach performs better than its competitors, which use features concerning the words of preceding segments, even when the labels are obtained automatically. According to those studies, accuracy is expected to drop around 2 percentage points when using automatic labels. However, in the context of a dialog system, the system is aware of the dialog acts of its own segments. Thus, only the classification of user segments is subject to error, which shall reduce the accuracy drop. Still, as future work, it is important to assess the concrete performance decay in this scenario.

Additionally, since the DIHANA corpus has hierarchical dialog act labels, when dealing with a certain level, we also explore the use of context information from the upper levels, relative to both the current and preceding segments. To provide this information we use the same label-based representation approach described for providing context information from the preceding segments.

\subsubsection{Dimensionality Reduction Layer}

In order to avoid result differences caused by using segment representations with different dimensionality, our architecture includes a dimensionality reduction layer that maps the generated segment representations into a 100-dimensional space. This way, the observed differences in performance are due to the nature of the segment representation approach and the information it is able to capture and not to factors related to the dimensionality. Furthermore, in order to reduce the probability of overfitting to the training data, this layer also applies dropout, disabling 50\% of the neurons during the training phase.

\subsubsection{Output Layer}

The output layer maps the 100-dimensional representation into a dialog act label. This is done using a dense layer with number of units equal to the number of labels. Since each segment has a single Level 1 label, we use the softmax activation together with the categorical cross entropy loss function when predicting those labels. However, that is not valid for the other levels, since they allow each segment to have multiple labels. Thus, in those cases, we use the sigmoid activation together with the binary cross entropy loss function, which, given the possibility of multiple labels, is actually the Hamming loss function, which is appropriate for this kind of problem~\citep{Diez2015}. In both cases, for performance reasons, we use the Adam optimizer~\citep{Kingma2015}.

%
%

%
%

\subsection{Training and Evaluation}
\label{ssec:eval}

To implement our networks we used Keras~\citep{Chollet2015} with the TensorFlow~\citep{Abadi2015} backend. We used mini-batching with batches of size 512 and the training phase stopped after 10 epochs without improvement on the validation set. Since there is some non-determinism involved, the results presented in the next section refer to the mean ($m$) and standard deviation ($s$) of the results obtained over 10 runs.

In order to evaluate our approaches, we performed 5-fold cross-validation using the folds defined in the first experiments on the DIHANA corpus~\citep{Tamarit2008,Martinez-Hinarejos2008}. The evaluation metrics vary according to the level of the hierarchy in focus. Since each segment has a single Level 1 label, at this stage we are dealing with a single-label classification problem. Thus, similarly to previous approaches on dialog act recognition, performance can be evaluated using accuracy. However, that is not the most appropriate metric for Levels 2 and 3, since they pose multi-label classification problems. Thus, we assess performance on those levels using the adapted metrics described by \citet{Sorower2010}. The multi-label equivalent of accuracy is the exact match ratio (MR), defined as

\begin{equation}
\label{eq:mr}
	\text{MR} = \frac{1}{n}\sum_{i=1}^{n}I(Y_i = Z_i),
\end{equation}

\noindent
where $Y_i$ is the set of gold standard labels of example $i$, $Z_i$ is the set of labels predicted by the classifier for the same example, and $I$ is the indicator function. The problem with this metric is that it does not account for partial correctness, which is common in multi-label classification problems. Thus, the single-label metrics of accuracy (Acc), precision (P), recall (R) and F-measure (F$_1$) are adapted to the multi-label problem as follows:

\begin{equation}
\label{eq:acc}
	\text{Acc} = \frac{1}{n}\sum_{i=1}^{n}\frac{|Y_i \cap Z_i|}{|Y_i \cup Z_i|},
\end{equation}

\begin{equation}
\label{eq:p}
	P = \frac{1}{n}\sum_{i=1}^{n}\frac{|Y_i \cap Z_i|}{|Z_i|},
\end{equation}

\begin{equation}
\label{eq:r}
	R = \frac{1}{n}\sum_{i=1}^{n}\frac{|Y_i \cap Z_i|}{|Y_i|},
\end{equation}

\begin{equation}
\label{eq:f}
	F_1 = \frac{1}{n}\sum_{i=1}^{n}\frac{2|Y_i \cap Z_i|}{|Y_i| + |Z_i|}.
\end{equation}

\noindent
where the operator $|X|$ is used to obtain the number of elements in the set $X$. Additionally, as previously stated, the Hamming loss (HL), which states how many times, on average, the relevance of an example to a class label is incorrectly predicted and is defined as

\begin{equation}
\label{eq:hl}
\begin{split}
	\text{HL} = \frac{1}{n|L|}\sum_{i=1}^{n}\sum_{l \in L} [I(l \in Z_i \land l \notin Y_i) + \\ + I(l \notin Z_i \land l \in Y_i)],
\end{split}
\end{equation}

\noindent
where $L$ is the set of all possible labels, is also an appropriate metric to evaluate the performance on multi-label classification problems. In the next section, the results for every metric except the Hamming loss are presented as a percentage.

To assess whether the differences between two approaches are statistically significant, we randomly selected one of the runs for each approach and performed a binomial test on their accuracy in Level 1 experiments and on their exact match ratio in Level 2 and 3 experiments. In the discussion in the next section we consider a confidence level of 95\%, that is, we consider that there is a statistically significant difference between the approaches if the $p$-value of the binomial test is lower than 0.05. 

%
%

%
%
%
%

\section{Results}
\label{sec:results}

Since each level in the hierarchical dialog act annotation of the DIHANA corpus has different characteristics and poses different problems, we start by presenting the results achieved on each of the levels independently. Furthermore, since we want to assess the importance of context information from upper levels, we start on the top level and descend the hierarchy. Finally, we present the results achieved on the hierarchical combination of the different levels.

\subsection{Level 1}

The results obtained when using the recurrent and convolutional segment representation approaches to predict Level 1 labels are shown in Table~\ref{tab:resl1}. We can see that the \ac{CNN}-based approach leads to better performance than the \ac{RNN}-based one ($p \approx 0.04$). However, both approaches lead to average accuracy results above 90\% and the difference between them is just 0.5 percentage points, which suggests that they are able to capture similar generic intention information. Still, while the network using the \ac{CNN}-based approach takes an average of 0.61 seconds per epoch to train and 27 epochs to converge, training the network using the \ac{RNN}-based approach takes much longer, with an average of 17.63 seconds per epoch and 46 epochs to converge. Additionally, as expected, using wider context windows around each token leads to better results ($p \approx 0.03$), which confirms that the generic Level 1 dialog acts are more related to the structure of the segment than to specific keywords. Still, since three different context windows are used in parallel and the two sets used in our experiments overlap, the accuracy difference between using the narrower windows used by \citet{Liu2017} in their study and the wider ones used by \citet{Kim2014} is just 0.24 percentage points. 

\begin{table} [ht]
\begin{center}
    \begin{tabular}{l c c}
        \toprule
        & \multicolumn{2}{c}{\textbf{Accuracy}} \tabularnewline
        \textbf{Approach} & $m$ & $s$ \tabularnewline
        \midrule
        Recurrent (RNN)               & 91.20 & 0.06 \tabularnewline
        Convolutional (CNN) w = [1,3] & 91.46 & 0.12 \tabularnewline
        Convolutional (CNN) w = [3,5] & 91.70 & 0.13 \tabularnewline
        \bottomrule
    \end{tabular}
\end{center}
\caption{Accuracy results on Level 1 using the two segment representation approaches.}
\label{tab:resl1}
\end{table}

Concerning context information provided by the preceding segments, the results in Table~\ref{tab:contextl1} show that the first preceding segment is the most important, leading to an accuracy improvement of 4.45 percentage points ($p \approx 6.7e^{-167}$). An additional improvement of 1.77 percentage points is achieved by providing information from two additional segments ($p \approx 4.6e^{-58}$). This pattern was expected, since it had already been observed in our study~\citep{Ribeiro2015} and that by \citet{Liu2017} on the \ac{SwDA} corpus, which is also annotated with task-independent dialog act labels.

\begin{table} [ht]
\begin{center}
    \begin{tabular}{l c c}
        \toprule
        & \multicolumn{2}{c}{\textbf{Accuracy}} \tabularnewline
        $n$ & $m$ & $s$ \tabularnewline
        \midrule
        0   & 91.70 & 0.13 \tabularnewline
        1   & 96.15 & 0.08 \tabularnewline
        2   & 97.47 & 0.06 \tabularnewline
        3   & 97.92 & 0.04 \tabularnewline
        \bottomrule
    \end{tabular}
\end{center}
\caption{Accuracy results on Level 1 using context information from $n$ preceding segments.}
\label{tab:contextl1}
\end{table}

\begin{table} [ht]
\begin{center}
    \begin{tabular}{l c c}
        \toprule
        & \multicolumn{2}{c}{\textbf{Accuracy}} \tabularnewline
        \textbf{Speaker} & $m$ & $s$ \tabularnewline
        \midrule
        User    & 95.17 & 0.12 \tabularnewline
        System  & 99.91 & 0.00 \tabularnewline
        \bottomrule
    \end{tabular}
\end{center}
\caption{Level 1 accuracy results on user and system segments.}
\label{tab:userl1}
\end{table}

When information from three preceding segments is used, the classifier only fails to accurately predict two percent of the segments. That result takes into account all the segments in the DIHANA Corpus. However, the system segments are scripted and, thus, are easier to predict than the user segments. In fact, if we consider the scenario of a dialog system trying to predict dialog acts, it is aware of its own and must only predict those of its conversational partners. In this sense, in Table~\ref{tab:userl1} we can see the results achieved when considering user and system segments independently. As expected, the average accuracy on system segments is 99.91\%. On user segments that value decreases to 95.17\%, which still reveals high performance.

Looking at each label individually, the hardest to identify is \textit{Undefined}, with a recall around 57\%. This was expected since that label covers all the cases which cannot be labeled with any of the other labels, including problems in the dialog. All the remaining labels have a recall above 95\%, with the lowest being that of the \textit{Answer} label, which is also the lowest in terms of precision (96\%). In both cases, the confusion is typically with the \textit{Question} label, which makes sense, since questions and answers may have the same words and only differ in terms of their order. In fact, if we consider questions in declarative form, there may be no difference at all.

Considering previous studies on dialog act recognition on the DIHANA Corpus, only \citet{Tamarit2008} assessed the performance on the Level 1 alone, achieving 60.70\% accuracy. However, their study focused on the use of prosodic information and, thus, it is not fair to compare their results with ours, since our approach takes advantage of the transcriptions.

\subsection{Level 2}

\begin{table*} [htpb]
\begin{center}
    \resizebox{\textwidth}{!}{
    \begin{tabular}{l c c c c c c c c c c c c}
        \toprule
        & \multicolumn{2}{c}{\textbf{MR}} & \multicolumn{2}{c}{\textbf{Acc}} & \multicolumn{2}{c}{\textbf{P}} & \multicolumn{2}{c}{\textbf{R}} & \multicolumn{2}{c}{\textbf{F$_1$}} & \multicolumn{2}{c}{\textbf{HL}} \tabularnewline
        \textbf{Approach} &   $m$ &  $s$ &   $m$ &  $s$ &   $m$ &  $s$ &   $m$ &  $s$ &   $m$ &  $s$ &    $m$ &    $s$ \tabularnewline
        \midrule
        RNN      	      & 69.65 & 0.50 & 70.42 & 0.48 & 71.10 & 0.46 & 70.51 & 0.47 & 70.68 & 0.47 & 0.0381 & 0.0004 \tabularnewline
        CNN w = [1,3]     & 70.71 & 0.33 & 71.58 & 0.33 & 72.30 & 0.33 & 71.74 & 0.34 & 71.87 & 0.33 & 0.0381 & 0.0002 \tabularnewline
        CNN w = [3,5]     & 70.24 & 0.27 & 71.17 & 0.26 & 71.93 & 0.28 & 71.33 & 0.26 & 71.48 & 0.26 & 0.0383 & 0.0000 \tabularnewline
        \bottomrule
    \end{tabular}
    }
    \caption{Results on Level 2 using the two segment representation approaches.}
    \label{tab:resl2}
    \bigskip
    \begin{tabular}{l c c c c c c c c c c c c}
        \toprule
        & \multicolumn{2}{c}{\textbf{MR}} & \multicolumn{2}{c}{\textbf{Acc}} & \multicolumn{2}{c}{\textbf{P}} & \multicolumn{2}{c}{\textbf{R}} & \multicolumn{2}{c}{\textbf{F$_1$}} & \multicolumn{2}{c}{\textbf{HL}} \tabularnewline
        $n$ &   $m$ &  $s$ &   $m$ &  $s$ &   $m$ &  $s$ &   $m$ &  $s$ &   $m$ &  $s$ &    $m$ &    $s$ \tabularnewline
        \midrule
        0   & 70.71 & 0.33 & 71.58 & 0.33 & 72.30 & 0.33 & 71.74 & 0.34 & 71.87 & 0.33 & 0.0381 & 0.0002 \tabularnewline
        1   & 91.07 & 0.14 & 91.52 & 0.13 & 91.84 & 0.13 & 91.67 & 0.13 & 91.68 & 0.13 & 0.0121 & 0.0002 \tabularnewline
        2   & 92.52 & 0.09 & 92.99 & 0.08 & 93.30 & 0.08 & 93.12 & 0.09 & 93.14 & 0.08 & 0.0101 & 0.0001 \tabularnewline
        3   & 92.97 & 0.12 & 93.45 & 0.11 & 93.75 & 0.09 & 93.61 & 0.11 & 93.60 & 0.10 & 0.0094 & 0.0001 \tabularnewline
        \bottomrule
    \end{tabular}
    \caption{Results on Level 2 using context information from $n$ preceding segments.}
    \label{tab:contextl2}
    \bigskip
    \begin{tabular}{l c c c c c c c c c c c c}
        \toprule
        & \multicolumn{2}{c}{\textbf{MR}} & \multicolumn{2}{c}{\textbf{Acc}} & \multicolumn{2}{c}{\textbf{P}} & \multicolumn{2}{c}{\textbf{R}} & \multicolumn{2}{c}{\textbf{F$_1$}} & \multicolumn{2}{c}{\textbf{HL}} \tabularnewline
        $n$ &   $m$ &  $s$ &   $m$ &  $s$ &   $m$ &  $s$ &   $m$ &  $s$ &   $m$ &  $s$ &    $m$ &    $s$ \tabularnewline
        \midrule
        0   & 93.18 & 0.18 & 93.68 & 0.16 & 93.99 & 0.15 & 93.87 & 0.15 & 93.63 & 0.15 & 0.0092 & 0.0002 \tabularnewline
        1   & 94.28 & 0.15 & 94.75 & 0.14 & 95.06 & 0.13 & 94.91 & 0.13 & 94.91 & 0.13 & 0.0077 & 0.0002 \tabularnewline
        2   & 94.29 & 0.05 & 94.76 & 0.05 & 95.06 & 0.05 & 94.91 & 0.06 & 94.91 & 0.06 & 0.0077 & 0.0001 \tabularnewline
        3   & 94.38 & 0.11 & 94.84 & 0.11 & 95.15 & 0.12 & 94.97 & 0.12 & 94.99 & 0.12 & 0.0075 & 0.0001 \tabularnewline
        \bottomrule
    \end{tabular}
    \caption{Results on Level 2 using Level 1 context information from $n$ preceding segments.}
    \label{tab:uppercontextl2}
    \bigskip
    \resizebox{\textwidth}{!}{
    \begin{tabular}{l c c c c c c c c c c c c}
        \toprule
        & \multicolumn{2}{c}{\textbf{MR}} & \multicolumn{2}{c}{\textbf{Acc}} & \multicolumn{2}{c}{\textbf{P}} & \multicolumn{2}{c}{\textbf{R}} & \multicolumn{2}{c}{\textbf{F$_1$}} & \multicolumn{2}{c}{\textbf{HL}} \tabularnewline
        \textbf{Speaker} &   $m$ &  $s$ &   $m$ &  $s$ &   $m$ &  $s$ &   $m$ &  $s$ &   $m$ &  $s$ &    $m$ &    $s$ \tabularnewline
        \midrule
        User             & 91.28 & 0.24 & 92.08 & 0.21 & 92.62 & 0.19 & 92.32 & 0.22 & 92.34 & 0.21 & 0.0115 & 0.0003 \tabularnewline
        System           & 98.43 & 0.09 & 98.44 & 0.08 & 98.44 & 0.08 & 98.45 & 0.07 & 98.44 & 0.08 & 0.0024 & 0.0001 \tabularnewline
        \bottomrule
    \end{tabular}
    }
    \caption{Level 2 results on user and system segments.}
    \label{tab:userl2}
\end{center}
\end{table*}

As stated in Section~\ref{ssec:dihana}, some Level 1 labels can only be paired with the \textit{Nil} label on the remaining levels. Thus, segments labeled with one of those labels on Level 1 have their labels on the remaining levels defined, independently of their content. Thus, we do not take those segments into account in our experiments on Levels 2 and 3.

Similarly to what happened on Level 1, in Table~\ref{tab:resl2}, we can see that using the \ac{CNN}-based segment representation approach leads to better results than the \ac{RNN}-based one. The only exception is the Hamming loss, which, on average, is equal for both approaches. On every non-loss metric, the \ac{CNN}-based approach surpasses the \ac{RNN}-based one by over 1 percentage point ($p \approx 1.1e^{-14}$). In this case, the discrepancy in the number of epochs required for training is smaller, with an average of 46 for the \ac{CNN}-based approach and 56 for the \ac{RNN}-based one. Furthermore, since we are considering less segments, the training times per epoch are reduced to 0.40 and 11.67 seconds, respectively.

Contrarily to what happened on Level 1, using narrower context windows apparently leads to better results. However, the difference is not statistically significant ($p \approx 0.12$). Still, this shows that task-specific dialog act labels are more related to certain keywords than the generic labels of Level 1. Furthermore, since the number of labels per segment is typically low, the classifiers tend to avoid selecting incorrect labels, which is reflected in higher precision than recall for every approach.

The results in Table~\ref{tab:contextl2} show that, similarly to what happened on Level 1, the preceding segments are able to provide relevant context information for the task. However, in this case, the importance of the first preceding segment is more pronounced, reducing the loss to less than a third and improving the remaining metrics by around 20 percentage points ($p \approx 5.0e^{-324}$. This makes sense considering that the dialogs feature many question-answer pairs focused on the same target information, which is the focus of Level 2 labels. Thus, in those cases, the labels of both segments are the same. Consequently, the labels of the first preceding segment provide an important cue for the identification of those of the current segment.

In Table~\ref{tab:uppercontextl2}, we can see that context information from Level 1 is also important. Using information from the current segment only leads to a slight but significant improvement ($p \approx 0.01$). However, also considering the Level 1 classification of the first preceding segment leads to an improvement around 1.5 percentage points on every non-loss metric ($p \approx 8.7e^{-6}$). This is still explained by the presence of multiple question-answer pairs in the dialogs, as if the preceding segment is labeled as \textit{Question} on Level 1, then the current segment probably has the same Level 2 labels as the preceding segment. The improvement achieved using information from additional preceding segments is not statistically significant ($p \approx 0.74$).

Similarly to what happened on Level 1, the performance on user segments is different from that on system segments. In Table~\ref{tab:userl2}, we can see that on system segments, the average value of every non-loss metric is around 98.4\%, while on user segments the average exact match ratio is 91.28\% and the remaining non-loss metrics are around 92\%.

Considering the labels individually, the best approach is unable to identify any of the three less predominant labels in the dataset. However, this was expected, since none of them appears in more than 29 segments. Thus, they are irrelevant for an approach focused on reducing the loss on the overall dataset and require specialized approaches or additional data to be identified. The \textit{Arrival Time} label has an F-measure around 75\% since it is easily confused with the \textit{Departure Time} label and is the less predominant of the two. Although the \textit{Train Type} label has precision above 95\%, it only has around 87\% recall. This happens since the label only appears in 2\% of the segments. Thus, in segments that focus on multiple aspects, information from the keywords that refer to the type of train is neglected in favor of that which allows the identification of more predominant labels. All the remaining labels have an F-measure above 95\% with balanced precision and recall. 

Previous studies on dialog act recognition on the DIHANA Corpus did not explore the Level 2 on its own, but rather combined it with the Level 1, using the combination of the labels of the two levels as the label set and looking at the problem as a single-label classification problem, similar to the classification of Level 1. Thus, our results on the Level 2 cannot be compared directly with those of previous studies. The results achieved on the combination of both levels are discussed in Section~\ref{ssec:hierarchical}.

\subsection{Level 3}

\begin{table*} [ht]
\begin{center}
    \resizebox{\textwidth}{!}{
    \begin{tabular}{l c c c c c c c c c c c c}
        \toprule
        & \multicolumn{2}{c}{\textbf{MR}} & \multicolumn{2}{c}{\textbf{Acc}} & \multicolumn{2}{c}{\textbf{P}} & \multicolumn{2}{c}{\textbf{R}} & \multicolumn{2}{c}{\textbf{F$_1$}} & \multicolumn{2}{c}{\textbf{HL}} \tabularnewline
        \textbf{Approach} &   $m$ &  $s$ &   $m$ &  $s$ &   $m$ &  $s$ &   $m$ &  $s$ &   $m$ &  $s$ &    $m$ &    $s$ \tabularnewline
        \midrule
        RNN               & 95.79 & 0.24 & 96.61 & 0.29 & 96.84 & 0.29 & 96.81 & 0.30 & 96.78 & 0.30 & 0.0043 & 0.0004 \tabularnewline
        CNN w = [1,3]     & 96.01 & 0.08 & 96.88 & 0.10 & 97.11 & 0.10 & 97.08 & 0.12 & 97.05 & 0.11 & 0.0040 & 0.0000 \tabularnewline
        CNN w = [3,5]     & 95.35 & 0.23 & 96.26 & 0.18 & 96.51 & 0.17 & 96.45 & 0.15 & 96.44 & 0.16 & 0.0046 & 0.0002 \tabularnewline
        \bottomrule
    \end{tabular}
    }
    \caption{Results on Level 3 using the two segment representation approaches.}
    \label{tab:resl3}
    \bigskip
    \begin{tabular}{l c c c c c c c c c c c c}
        \toprule
        & \multicolumn{2}{c}{\textbf{MR}} & \multicolumn{2}{c}{\textbf{Acc}} & \multicolumn{2}{c}{\textbf{P}} & \multicolumn{2}{c}{\textbf{R}} & \multicolumn{2}{c}{\textbf{F$_1$}} & \multicolumn{2}{c}{\textbf{HL}} \tabularnewline
        $n$ &   $m$ &  $s$ &   $m$ &  $s$ &   $m$ &  $s$ &   $m$ &  $s$ &   $m$ &  $s$ &    $m$ &    $s$ \tabularnewline
        \midrule
        0   & 96.01 & 0.08 & 96.88 & 0.10 & 97.11 & 0.10 & 97.08 & 0.12 & 97.05 & 0.11 & 0.0040 & 0.0000 \tabularnewline
        1   & 96.05 & 0.13 & 96.91 & 0.10 & 97.14 & 0.09 & 97.10 & 0.09 & 97.08 & 0.10 & 0.0039 & 0.0001 \tabularnewline
        2   & 96.10 & 0.16 & 96.95 & 0.11 & 97.17 & 0.11 & 97.14 & 0.10 & 97.12 & 0.10 & 0.0039 & 0.0002 \tabularnewline
        3   & 96.10 & 0.16 & 96.96 & 0.13 & 97.19 & 0.13 & 97.14 & 0.11 & 97.13 & 0.12 & 0.0039 & 0.0001 \tabularnewline
        \bottomrule
    \end{tabular}
    \caption{Results on Level 3 using context information from $n$ preceding segments.}
    \label{tab:contextl3}
    \bigskip
    \begin{tabular}{l c c c c c c c c c c c c}
        \toprule
        & \multicolumn{2}{c}{\textbf{MR}} & \multicolumn{2}{c}{\textbf{Acc}} & \multicolumn{2}{c}{\textbf{P}} & \multicolumn{2}{c}{\textbf{R}} & \multicolumn{2}{c}{\textbf{F$_1$}} & \multicolumn{2}{c}{\textbf{HL}} \tabularnewline
        $n$ &   $m$ &  $s$ &   $m$ &  $s$ &   $m$ &  $s$ &   $m$ &  $s$ &   $m$ &  $s$ &    $m$ &    $s$ \tabularnewline
        \midrule
        0   & 96.20 & 0.15 & 97.03 & 0.11 & 97.25 & 0.11 & 97.20 & 0.10 & 97.19 & 0.11 & 0.0037 & 0.0002 \tabularnewline
        1   & 96.24 & 0.09 & 97.05 & 0.08 & 97.28 & 0.08 & 97.22 & 0.07 & 97.21 & 0.08 & 0.0037 & 0.0000 \tabularnewline
        2   & 96.29 & 0.08 & 97.11 & 0.08 & 97.34 & 0.09 & 97.27 & 0.09 & 97.26 & 0.09 & 0.0036 & 0.0000 \tabularnewline
        3   & 96.17 & 0.06 & 97.00 & 0.06 & 97.23 & 0.06 & 97.18 & 0.06 & 97.17 & 0.06 & 0.0038 & 0.0000 \tabularnewline
        \bottomrule
    \end{tabular}
    \caption{Results on Level 3 using Level 2 context information from $n$ preceding segments.}
    \label{tab:uppercontextl3-2}
    \bigskip    
    \begin{tabular}{l c c c c c c c c c c c c}
        \toprule
        & \multicolumn{2}{c}{\textbf{MR}} & \multicolumn{2}{c}{\textbf{Acc}} & \multicolumn{2}{c}{\textbf{P}} & \multicolumn{2}{c}{\textbf{R}} & \multicolumn{2}{c}{\textbf{F$_1$}} & \multicolumn{2}{c}{\textbf{HL}} \tabularnewline
        $n$ &   $m$ &  $s$ &   $m$ &  $s$ &   $m$ &  $s$ &   $m$ &  $s$ &   $m$ &  $s$ &    $m$ &    $s$ \tabularnewline
        \midrule
        0   & 96.32 & 0.12 & 97.13 & 0.10 & 97.36 & 0.09 & 97.29 & 0.10 & 97.28 & 0.09 & 0.0036 & 0.0001 \tabularnewline
        1   & 96.29 & 0.14 & 97.10 & 0.12 & 97.33 & 0.12 & 97.26 & 0.11 & 97.26 & 0.12 & 0.0037 & 0.0001 \tabularnewline
        2   & 96.34 & 0.12 & 97.14 & 0.11 & 97.36 & 0.10 & 97.30 & 0.11 & 97.30 & 0.11 & 0.0036 & 0.0001 \tabularnewline
        3   & 96.30 & 0.13 & 97.13 & 0.11 & 97.35 & 0.10 & 97.31 & 0.10 & 97.29 & 0.10 & 0.0037 & 0.0001 \tabularnewline
        \bottomrule
    \end{tabular}
    \caption{Results on Level 3 using Level 1 context information from $n$ preceding segments.}
    \label{tab:uppercontextl3-1}
    \bigskip
    \resizebox{\textwidth}{!}{
    \begin{tabular}{l c c c c c c c c c c c c}
        \toprule
        & \multicolumn{2}{c}{\textbf{MR}} & \multicolumn{2}{c}{\textbf{Acc}} & \multicolumn{2}{c}{\textbf{P}} & \multicolumn{2}{c}{\textbf{R}} & \multicolumn{2}{c}{\textbf{F$_1$}} & \multicolumn{2}{c}{\textbf{HL}} \tabularnewline
        \textbf{Speaker} &   $m$ &  $s$ &   $m$ &  $s$ &   $m$ &  $s$ &   $m$ &  $s$ &   $m$ &  $s$ &    $m$ &    $s$ \tabularnewline
        \midrule
        User             & 95.58 & 0.16 & 95.62 & 0.17 & 95.62 & 0.17 & 95.65 & 0.18 & 95.63 & 0.17 & 0.0044 & 0.0002 \tabularnewline
        System           & 97.55 & 0.12 & 99.06 & 0.06 & 99.52 & 0.06 & 99.25 & 0.03 & 99.33 & 0.04 & 0.0024 & 0.0001 \tabularnewline
        \bottomrule
    \end{tabular}
    }
    \caption{Level 3 results on user and system segments.}
    \label{tab:userl3}
\end{center}
\end{table*}

Table~\ref{tab:resl3} shows that, similarly to what happened on the other levels, the \ac{CNN}-based segment representation approach leads to better results than the \ac{RNN}-based one ($p \approx 9.6e^{-5}$). However, in this case, the difference is less pronounced. In fact, when using the set of wider windows, the \ac{CNN}-based approach performs worse than the \ac{RNN}-based one ($p \approx 1.2e^{-4}$). This is due to the fact that the Level 3 focuses on the information that is explicitly referred to in the segments and, thus, is even more keyword-oriented than Level 2. That also explains the average results above 96\% on every non-loss metric. The average training times per epoch are the same as those for Level 2. However, in this case, more epochs are required until convergence {---} 86 for the \ac{RNN}-based approach and 80 for the \ac{CNN}-based one.

The results in Table~\ref{tab:contextl3} show that, in this case, the improvement provided by context information from the preceding segments is negligible and not statistically significant ($p \approx 0.48$). Once again, this is explained by the nature of Level 3 and its focus on what is explicitly referred to in the current segment. Thus, the preceding segments are not relevant.

In Table~\ref{tab:uppercontextl3-2} we can see that the improvement provided by Level 2 information is slightly superior than that provided by the Level 3 information from the preceding segments. In this case, considering the Level 2 label of the same segment leads to a statistically significant improvement ($p \approx 0.03$). This can be explained by the fact that when a certain kind of information is explicitly referred to in a segment, it is typically also focused by the segment and, thus, overlaps between the Level 2 and 3 labels of a segment are common. Considering Level 2 information from preceding segments does not lead to statistically significant improvements ($p \approx 0.13$).

Since the Level 1 labels are related to the generic intention of the segment, they have no direct relation to what is explicitly referred to in the segment and, thus, to the Level 3. This is confirmed by the results in Table~\ref{tab:uppercontextl3-1}, which show that the improvement provided by Level 1 information is negligible and not statistically significant ($p \approx 0.13$). 

In Table~\ref{tab:userl3}, we can see that, in this case, the performance difference between user and system segments is not as pronounced. Once again, this is explained by the fact that the Level 3 is highly focused on keywords and, thus, the fact that the system segments are scripted does not have the same influence on classification. 

Considering the labels individually, similarly to what happened on Level 2, the best approach is unable to identify the less predominant labels, \textit{Duration} and \textit{Service}, since none of them appears in more than 19 segments. Of the remaining labels, \textit{Arrival Time} is that with lowest recall, 88\%, since it is easily confused with the more predominant \textit{Departure Time} label. All the remaining labels have an F-measure above 97\% with balanced precision and recall. 

Similarly to the Level 2, previous studies on dialog act recognition on the DIHANA corpus did not explore the Level 3 alone, but rather combined it with the remaining levels. Consequently, we are also unable to directly compare our results on the Level 3 with those of previous studies. The hierarchical combination of the multiple levels is explored in the next section.

\subsection{Hierarchical Classification}
\label{ssec:hierarchical}

As previously stated, previous studies on dialog act recognition on the DIHANA corpus did not explore the task-specific levels of the hierarchy independently, but rather in combination with the levels above them. This makes sense from a hierarchical point of view, as each level is supposed to depend on those above it. However, as discussed in Section~\ref{ssec:dihana}, since each level focuses on a different aspect concerning the intention of the speaker, the only restriction imposed by the annotation scheme is that segments annotated with a Level 1 label that refers to dialog structuring or communication problems cannot have labels on the remaining levels. Still, the results reported in the previous sections show that the ability to predict the label at a given level is improved when context information from the level directly above it is used. Furthermore, in order to accurately identify the intention of a speaker, the system must be able to accurately predict the labels at the three levels. Thus, we also assess the performance on the hierarchical combination of the multiple levels.

The previous studies on the task approached the problem of the combined classification of the different levels as a single-label classification problem in which each combination of labels present in the corpus is considered a single independent label. However, this approach has two flaws. On the one hand, it is a simplification of the problem as it limits the possible labels to the combinations existing in the dataset. On the other hand, it does not take the multi-label nature of the task-specific levels into account.

Contrarily to those studies, we approach the problem hierarchically by combining the best classifiers for each level. That is, for each segment, we start by predicting its Level 1 label using the \ac{CNN}-based classifier with wide context windows and context information from three preceding segments. Then, we predict its Level 2 labels using the \ac{CNN}-based classifier with narrow context windows, Level 2 context information from three preceding segments, and Level 1 context information from the current and first preceding segment. Finally, we predict its Level 3 labels using the \ac{CNN}-based classifier with narrow context windows and Level 2 context information from the current segment. In order to account for the fact that the Level 2 and 3 classifiers were not trained on the segments with Level 1 labels that do not allow labels on the remaining levels, if the Level 1 classifier predicts one of those labels for the segment, it is automatically assigned no labels on the remaining levels.

Using this hierarchical approach, the bottom levels are still considered multi-label classification problems. Thus, every combination of labels is possible and not just those that appear on the dataset. Still, in order to confirm that the problem approached by previous studies is actually simpler, we also present the results achieved when the task is approached as a single-label classification problem. To obtained those results, we used a classifier with the same architecture as the best Level 1 classifier, that is, a \ac{CNN}-based classifier with wide context windows and context information from three preceding segments. However, in this case, the classifier was trained to predict the combination of all labels for the segment at once and each of those combinations is seen as an independent label. 

For comparison with the results achieved on previous studies, we use the exact match ratio to evaluate the performance of both the hierarchical and single-label approaches. Thus, if the prediction of the Level 1 label is inaccurate or there is any missing or additional Level 2 or 3 label, the whole prediction for the segment is considered wrong. 

Table~\ref{tab:resl12} shows the results achieved on the combination of Levels 1 and 2. Using the hierarchical approach we achieved an average of 94.28\% exact match ratio, which is already above the 93.40\% reported by \citet{Martinez-Hinarejos2008} ($p \approx 3.0e^{-8}$) and in line with the 94.08\% reported by \citet{Gamback2011} ($p \approx 0.20$). By approaching the task as a single-label classification problem we achieved 96.24\%, which is almost two percentage points above the result achieved using the hierarchical approach ($p \approx 7.0e^{-43}$). This confirms that this view on the problem is actually a simplification. 

\begin{table} [ht]
\begin{center}
    \resizebox{\columnwidth}{!}{
    \begin{tabular}{l c c}
        \toprule
        & \multicolumn{2}{c}{\textbf{MR}} \tabularnewline
        \textbf{Approach} &   $m$ &  $s$ \tabularnewline
        \midrule     
        Hierarchical      & 94.28 & 0.03 \tabularnewline
        Single-Label      & 96.24 & 0.06 \tabularnewline
        \midrule
        \citet{Martinez-Hinarejos2008} & \multicolumn{2}{c}{93.40}\tabularnewline
        \citet{Gamback2011}            & \multicolumn{2}{c}{94.08}\tabularnewline
        \bottomrule
    \end{tabular}
    }
\end{center}
\caption{Results achieved on the combination of Levels 1 and 2.}
\label{tab:resl12}
\end{table}

Table~\ref{tab:resl123} shows the results achieved on the combination of the three levels. We can see that most of the conclusions drawn for the combination of Levels 1 and 2 can also be drawn in this case. Using the hierarchical approach we achieved an average of 92.34\% exact match ratio, which is above the 89.70\% reported by \citet{Martinez-Hinarejos2008} ($p \approx 9.5e^{-44}$) and the 90.97\% reported by \citet{Gamback2011} . However, while on the combination of the two top levels the result of the hierarchical approach was not statistically different from that reported by \citet{Gamback2011}, in this case there is a statistically significant improvement of 1.37 percentage points ($p \approx 6.6e^{-14}$). By approaching the task as a single-label classification problem, the exact match ratio is improved to 93.98\% ($p \approx 1.5e^{-22}$), once again confirming that the problem is simpler.

\begin{table} [ht]
\begin{center}
    \resizebox{\columnwidth}{!}{
    \begin{tabular}{l c c}
        \toprule
        & \multicolumn{2}{c}{\textbf{MR}} \tabularnewline
        \textbf{Approach} &   $m$ &  $s$ \tabularnewline
        \midrule
        Hierarchical      & 92.34 & 0.04 \tabularnewline
        Single-Label      & 93.98 & 0.19 \tabularnewline
        \midrule
        \citet{Martinez-Hinarejos2008} & \multicolumn{2}{c}{89.70}\tabularnewline
        \citet{Gamback2011}            & \multicolumn{2}{c}{90.97}\tabularnewline
        \bottomrule
    \end{tabular}
    }
\end{center}
\caption{Results achieved on the combination of all levels.}
\label{tab:resl123}
\end{table}

%
%

%
%

\section{Conclusions}
\label{sec:conclusions}

In this article we have explored automatic dialog act recognition on the DIHANA corpus. This dataset and its three-level annotation scheme pose problems which have been neglected since the studies on dialog act recognition started focusing on English data and, especially, on the \ac{SwDA} corpus. The first problem concerns the language difference. Additionally, contrarily to the flat and single-label classification problem posed by the SWBD-DAMSL annotations of the \ac{SwDA} corpus, the dialog act annotations of the DIHANA corpus pose a hierarchical classification problem. Furthermore, the two lower levels of that hierarchy pose multi-label classification problems. We have studied how the state-of-the-art approaches on dialog act recognition on English data can be applied to these problems and which aspects of those approaches are relevant for the prediction of the labels of each level, according to its characteristics.

A conclusion that was common to all levels was that the \ac{CNN}-based approach on segment representation led to better performance than the \ac{RNN}-based approach. This approach, applied to dialog act recognition on English data by \citet{Liu2017}, features three parallel temporal \acp{CNN} with context windows of different sizes. This way, the segment representation approach takes sets of words of different sizes into account and, depending on the sizes of the windows, is able to capture information concerning both specific words and the structure of the segment. In this sense, the task-independent labels of Level 1 are more related to the structure of the segment and, thus, the best results were achieved using a set of wider context windows. On the other hand, the task-specific labels of Levels 2 and 3 are more related to certain keywords and, thus, using a set of narrower windows led to improved performance. The importance of the selected window sizes was especially pronounced on the experiments on Level 3, since when using wider windows the \ac{CNN}-based approach performed worse than the \ac{RNN}-based one. However, that is explainable by the nature of that level, which focuses on the kind of information that is explicitly referred to in the segments and, thus, the classification of a segment is given by the presence of specific words.

The relation between the Level 3 labels and the presence of certain keywords in the segment also explains the lack of importance of context information from the preceding segments to the prediction of those labels. On the other hand, that information is relevant for predicting the labels of the remaining levels. On Level 1, the experiments revealed a pattern similar to that revealed in both our study~\citep{Ribeiro2015} and that by \citet{Liu2017} on \ac{SwDA}, which is also annotated with task-independent labels. However, the importance of context information from the preceding segments was especially pronounced on the experiments on Level 2, reducing Hamming loss to less than a third and improving the remaining metrics by over 20 percentage points. The Level 2 focuses on the kind of information implicitly focused by the segment. Thus, since the dialogs in the DIHANA corpus feature multiple pairs of segments focused on the same kind of information, the preceding segments, especially the first, provide an important cue for the classification of the current segment.

Still considering the Level 2 and the characteristics of the dialogs, most of the pairs of segments focused on the same kind of information are question-answer pairs. \textit{Question} and \textit{Answer} are Level 1 labels. Thus, Level 1 context information from both the current and preceding segments also provides cues for the prediction of Level 2 labels. On the other hand, it is irrelevant when predicting Level 3 labels. However, there is a relation between the kind of information that is implicitly focused in a segment and that which is explicitly referred to in it. Thus, the sets of Level 2 and 3 labels of a segment typically overlap. Consequently, Level 2 context information is able to slightly improve the performance when predicting Level 3 labels.

The system segments of the DIHANA corpus are scripted and, thus, are easier to predict than the user segments. Furthermore, a dialog system is aware of the dialog acts of its own segments and must only predict those of its conversational partner's segments. Thus, for such a scenario, only the performance on user segments is relevant. As expected, the performance was higher on system segments on every level. However, on user segments, the average accuracy on Level 1 and the average exact match ratio on the remaining levels was still above 90\%. Furthermore, it is important to refer that since the Level 3 is highly keyword related, the performance difference is not as pronounced.

Finally, by hierarchically combining the best classifiers for each level, we achieved an average exact match ratio of 94.28\% on the combinations of Levels 1 and 2 and 92.34\% on the combination of the three levels. These results are already in line or above those achieved on previous studies on dialog act recognition on the DIHANA corpus. However, those studies considered a simplified version of the problem by reducing it to a single-label classification problem with the label of a segment consisting of the concatenation of the labels of the three levels. Since this approach only considers the label combinations present in the corpus, the number of possible labels is reduced in comparison to our approach, which looks at the prediction of Level 2 and 3 labels as multi-label classification problems. By approaching the problem in a manner comparable to that of those studies, the previous values increase to 96.24\% and 93.98\%, respectively.

In terms of future work it would be interesting to assess whether the conclusions drawn on this study on Spanish data and previously on English data also hold on data in other languages with different morphological typology. In terms of multi-label dialog act recognition, it would be interesting to explore the use of other loss functions when training the network, especially one based on F-measure, which is not as influenced by the reduced number of positive classes per segment as the Hamming loss. Furthermore, it is important to assess whether segment representation approaches based on character-level tokenization are able to capture additional information for predicting the task-specific labels. It would also be interesting to explore means to perform the hierarchical classification of the multiple levels using a single network instead of three independent classifiers. Finally, it is important to assess the decay in performance in a real scenario. That is, one in which the dialog system is not simulated and, thus, must deal with problems related to \ac{ASR} and use automatically predicted labels as context information.

%
%

%
%

\section*{Acknowledgements}
\label{sec:acknowledgements}

This work was supported by national funds through \ac{FCT} with reference UID/CEC/50021/2019 and by Universidade de Lisboa.

We would like to thank the editorial board of Linguamática for allowing us to distribute this translated version of the article. When referring to this study, please cite its original version in Portuguese~\citep{Ribeiro2019}.

%
%

\bibliography{references}

\end{document}